\documentclass{article}

\usepackage{PRIMEarxiv}

\usepackage[utf8]{inputenc} % allow utf-8 input
\usepackage[T1]{fontenc}    % use 8-bit T1 fonts
\usepackage{hyperref}       % hyperlinks
\usepackage{url}            % simple URL typesetting
\usepackage{booktabs}       % professional-quality tables
\usepackage{amsfonts}       % blackboard math symbols
\usepackage{nicefrac}       % compact symbols for 1/2, etc.
\usepackage{microtype}      % microtypography
\usepackage{lipsum}
\usepackage{fancyhdr}       % header
\usepackage{graphicx}       % graphics
\graphicspath{{media/}}     % organize your images and other figures under media/ folder

\usepackage{CJKutf8}
\usepackage{amsmath}
\usepackage{color}
\usepackage{enumerate}
\usepackage{algorithm, algorithmic}
\usepackage{multirow}

\title{Research on Domain-Specific Chinese Spelling Correction Method Based on Plugin Extension Modules
}

\author{
  Xiaowu Zhang \\
 University of Science and Technology Beijing \\
  Beijing\\
  \texttt{zhangxw21@outlook.com} \\
   \And
  Hongfei Zhao\thanks{\;\;Corresponding authors}  \\
  Fudan University \\
  Shanghai\\
  \texttt{iioSnail@163.com} \\
    \And
  Xuan Chang \\
  Beijing DeepGlint Technology Co., Ltd \\
  Beijing\\
 \texttt{xuanchang@deepglint.com} \\
}

\begin{document}
\begin{CJK}{UTF8}{gbsn}
\maketitle
\begin{abstract}
This paper proposes a Chinese spelling correction method based on plugin extension modules, aimed at addressing the limitations of existing models in handling domain-specific texts. Traditional Chinese spelling correction models are typically trained on general-domain datasets, resulting in poor performance when encountering specialized terminology in domain-specific texts. To address this issue, we design an extension module that learns the features of domain-specific terminology, thereby enhancing the model's correction capabilities within specific domains. This extension module can provide domain knowledge to the model without compromising its general spelling correction performance, thus improving its accuracy in specialized fields. Experimental results demonstrate that after integrating extension modules for medical, legal, and official document domains, the model's correction performance is significantly improved compared to the baseline model without any extension modules.
\end{abstract}
% keywords can be removed

\section{Introduction}
Existing Chinese spelling correction datasets are mostly focused on open-domain content, such as Wang271K \cite{wana}, SIGHAN \cite{tsenn}, and CSCD-IME \cite{huuu}. Models trained on these datasets typically demonstrate strong correction capabilities on general-domain Chinese texts. However, in real-world applications, texts that require correction often originate from specific domains and contain a substantial amount of domain-specific terminology. Due to the lack of these specialized terms in training datasets, current models generally struggle to perform accurate corrections in such cases.

To address this issue, some research efforts have designed specific correction solutions tailored to different domains. These existing methods can be broadly categorized into two approaches:

The first approach involves building correction systems by targeting unique features of the desired domain. For instance, \cite{jiang4} introduced the MCSCSet dataset for the medical question-answering domain and developed a domain-specific spelling correction model, MedSpell. Although this approach shows good performance within its target domain, it has limited transferability to other domains.

The second approach focuses on fine-tuning existing open-domain models using domain-specific spelling correction datasets to enhance their performance in the target domain. For example, Hongqiu Wu et al. proposed the LEMON dataset \cite{wu5}, covering six distinct domains, and explored the performance of existing correction models on this dataset. \cite{Lv6} introduced the ECSpell model, applying it to datasets from three different domains. While this approach extends model capabilities across multiple domains, it requires collecting dedicated correction datasets and retraining the model for each target domain. Additionally, this method often faces the issue of "catastrophic forgetting," where the model loses its original correction capabilities for open-domain texts after fine-tuning on a specific domain.

To overcome the limitations of existing models in correcting domain-specific Chinese texts, this paper proposes a plugin-based extension module that enhances the model's vocabulary, enabling it to correct domain-specific terminology while preserving its original correction capabilities. The proposed method is compatible with various neural network-based correction models. Expanding to new domains only requires providing a list of domain-specific terms and performing a small amount of fine-tuning on the extension module.

Furthermore, to address the common issue of over-correction in existing models, we propose a dual-prediction change analysis algorithm. This algorithm performs two rounds of predictions and analyzes the changes in prediction results for each character to determine whether it has been over-corrected, thereby reducing the likelihood of over-corrections.

\section{Related Work}
Wang et al. proposed a method using the Fusion Lattice LSTM model for Chinese spelling correction \cite{wang7}. Xingyi Cheng and colleagues significantly advanced Chinese spelling correction by developing the SpellGCN model based on Graph Neural Networks \cite{cheng8}. This method first encodes sentences using a word embedding layer; the encoded vectors are then simultaneously fed into a BERT model for correction and into a Graph Neural Network for lookup. Chong Li et al. introduced two general approaches to enhance Chinese spelling correction models \cite{li9}. Yinghui Li and co-authors incorporated contrastive learning into the training framework for Chinese spelling correction models, introducing an Error-driven Contrastive Probability Optimization (ECOPO) strategy, which successfully improved the correction performance of multiple models \cite{li10}. \cite{zhang11} proposed a Self-Distillation Contrastive Learning (SDCL) architecture for Chinese spelling correction. \cite{sun12} introduced an Error-Guided Correction Model (EGCM).

\section{Method}

\begin{figure*}[htb]
  \centering
  \includegraphics[width=0.7\textwidth]{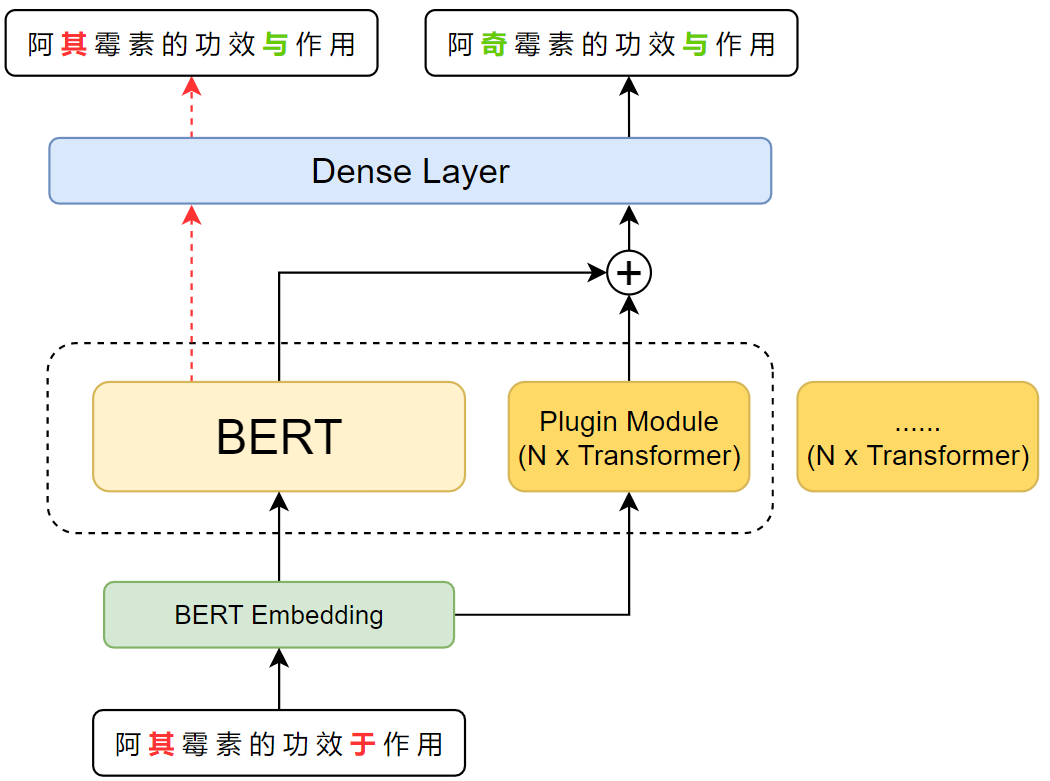}
  \caption[]{The main figure of the framework illustrates the overall structure of the domain-specific Chinese spelling correction method based on plugin extension modules. It details how the core modules interact with domain-specific knowledge bases or models through extension interfaces to achieve precise error correction and depicts the workflow from input to output, highlighting the logical connections between components.}
  \label{fig1}
\end{figure*}

The plugin-based extension model enhances the original model's ability to handle domain-specific terminology by incorporating an extension module that supplements the feature information from specific domains. The network architecture is shown in Figure 3-4. On the left side of the figure is the original model trained on open-domain datasets, which has strong correction capabilities for general Chinese texts. The original model comprises three main components: an embedding module, a feature extraction module, and a correction module. The embedding module maps Chinese text into 768-dimensional vectors for subsequent computations. The feature extraction module encodes each character using contextual information to refine its semantic representation, ensuring the accuracy of text features. Finally, the correction module maps the refined feature vectors to the indices of corresponding characters. However, since the model has not been exposed to domain-specific terminology during training, the feature extraction module struggles to encode these terms correctly, limiting the model’s correction performance. The forward propagation process of the original model is described by Equation \ref{eq1}:

\begin{equation}
    \label{eq1}
    Y = O(\text{BERT}(E(X)))
\end{equation}

where $X$ represents the input text, $E$ is the embedding module, $BERT$ is the feature extraction module, $O$ is the correction module, and $Y$ is the model output.

To enhance the model’s ability to encode features of specialized terminology, we introduce an extension module. This extension module consists of multiple layers of Transformer encoders, directly adopting the 12-layer Transformer architecture of BERT. The number of extension modules can be flexibly configured; however, when performing corrections, it is crucial to select an appropriate extension module. After the sentence is input into the model, it is first encoded by the embedding module, and then the encoded vectors are sent separately to the feature extraction module and the extension module for feature extraction. The two extracted feature sets are fused by addition, and the correction module then outputs the final correction result. The forward propagation process with the extension module is shown in Equation \ref{eq2}:

\begin{equation}
    \label{eq2}
    Y' = O(\text{BERT}(E(X)) + P(E(X)))
\end{equation}

where $P$ denotes the extension module.

\subsection{Unsupervised Training of the Extension Module}
The extension module is introduced to improve the model's correction capabilities for domain-specific terminology by providing specialized feature information to assist the correction module in making more precise corrections. When training the extension module, two key issues must be addressed. First, it is essential to avoid catastrophic forgetting, where the extension module interferes excessively with the feature vectors extracted by the original model, thereby ensuring that the model's correction performance on open-domain text remains unaffected. Second, due to the difficulty in collecting domain-specific sentence samples, the extension module must be capable of extracting effective feature information based solely on terminology.

To mitigate the issue of catastrophic forgetting, the training of the extension module freezes the parameters of the three modules in the original model, updating only the weights of the extension module. This approach offers two significant advantages: on one hand, not updating the parameters of the original model prevents any degradation in its correction capabilities, allowing it to retain its performance on open-domain texts when the extension module is not utilized. On the other hand, by fixing the original model's parameters, the learning of domain-specific features is entirely handled by the extension module, enhancing the model's modular expansion capabilities.

However, fixing the original model parameters introduces certain training challenges. During backpropagation, gradients are initially propagated to the parts closer to the correction module, then gradually to the feature extraction and extension modules. If the upstream correction module is not updated, the training effectiveness of the extension module may be diminished, but this issue can be alleviated by increasing the number of training iterations.

The training set for the extension module is derived from specialized terminology in the target domain, which can be sourced from public resources such as QQ Input Method Dictionary \footnote{http://cdict.qq.pinyin.cn/} and Sogou Cell Dictionary \footnote{https://pinyin.sogou.com/dict/}. These platforms' official and user dictionaries can be converted into text files, serving as the foundation for training the extension module. Constructing the training data involves two critical steps. The first step uses confusion sets to partially replace characters within the vocabulary. Specifically, 20\% of the terms remain unchanged, while 80\% are subjected to confusion: one character in a word is randomly replaced with a 30\% chance of replacing it with “[MASK]”, a 50\% chance of substituting it with a similar character from the confusion set, and a 20\% chance of replacing it with a random character. After constructing the input and label pairs, the second step adds identical characters randomly before and after the input and label sequences. For example, the sample “(\textcolor{red}{机其}学习, \textcolor{blue}{机器}学习)” is transformed into “(友哂摧屉蹂\textcolor{red}{机其}学习妩崩疲腠, 友哂摧屉蹂\textcolor{blue}{机器}学习妩崩疲腠)”. This operation introduces contextual noise, allowing the extension module’s self-attention mechanism to better capture the features of target terminology while ignoring non-target terms that should remain unchanged. This approach ensures that multiple erroneous forms of the same term can be learned during different inputs, while the labels remain consistent, thereby improving the model's correction capabilities.

\subsection{Dual-Prediction Change Analysis Strategy}

\begin{figure*}[ht]
  \centering
  \includegraphics[width=0.6\linewidth]{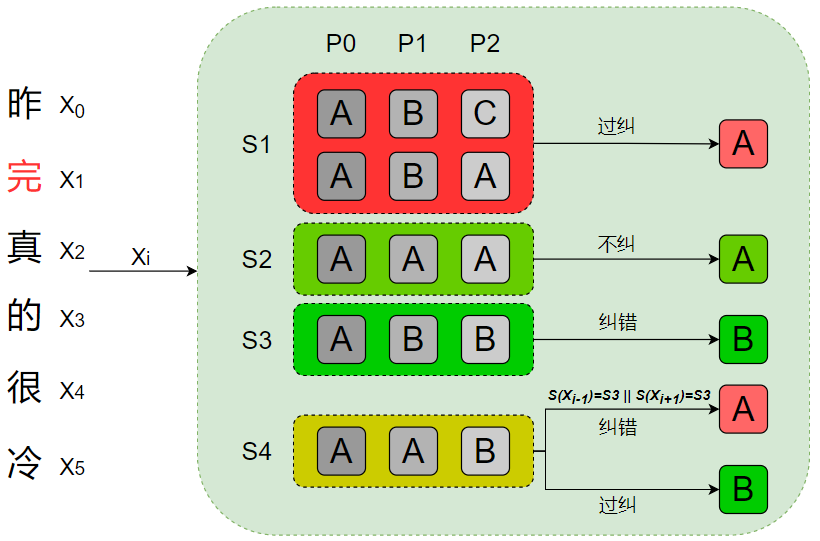}
  \caption[]{Dual-Prediction Change Analysis Algorithm Process}
  \label{fig2}
\end{figure*} 

Over-correction is a common issue in Chinese spelling correction models, where correct characters are mistakenly modified, reducing precision and affecting overall performance. To address this, we propose a dual-prediction change analysis algorithm that compares the results of two prediction rounds to determine whether each character has been over-corrected. The workflow is illustrated in Figure \ref{fig2}, which describes how the algorithm analyzes changes for a given character $x_i$. In the figure, $P0$ represents the original character $x_i$, $P1$ is the result from the first prediction, and $P2$ is the result from the second prediction. Five potential change scenarios are considered: “A→B→C”, “A→B→A”, “A→A→A”, “A→B→B”, and “A→A→B”, where “A, B, C” denote arbitrary characters. Based on these changes, the algorithm classifies each character into one of the following four cases:

\begin{enumerate}[a)]
    \item \textbf{Over-correction (S1):} If the model's two predictions for $x_i$ are inconsistent, indicating oscillation, the character is deemed over-corrected, and the original character $x_i$ is retained as the final output.
    \item \textbf{No correction needed (S2):} If neither prediction modifies $x_i$, no correction is required.
    \item \textbf{Confident correction (S3):} If both predictions consistently modify $x_i$ to the same character “B”, it indicates high confidence, and “B” is adopted as the final output.
    \item \textbf{Conditional correction (S4):} If the first prediction leaves $x_i$ unchanged, but the second modifies it to “B”, this may indicate either over-correction or a valid correction. A decision is made based on the surrounding characters $x_{i-1}$ and $x_{i+1}$; if both are classified as “S3”, it is considered a valid correction, otherwise, it is deemed over-correction.
\end{enumerate}

The dual-prediction change analysis algorithm operates as algorithm \ref{fun1}: first, the input sequence $X=\{x_0, x_1, \cdots, x_n\}$ is processed by the model $\textbf{F}$ for an initial prediction, yielding $Y^1=\{Y^1_0, Y^1_1, \cdots, Y^1_n\}$. Then, using $Y^1$ as the input, a second prediction produces $Y^2$. The algorithm then iterates through each character, recording its change status, and applies over-correction adjustments for cases “S1” and “S4”. For characters classified as “S1”, the original input is restored; for “S4” cases, the neighboring characters are examined to determine if over-correction occurred.

By analyzing the results from two prediction rounds, the dual-prediction change analysis algorithm effectively identifies and rectifies over-corrected characters, enhancing model precision without significantly compromising recall, thereby improving overall correction performance.

\begin{algorithm}[ht]
    \caption{Dual-Prediction Change Analysis Algorithm}
    \label{fun1}
    \begin{algorithmic}[1]
        \REQUIRE $X=\{x_0, x_1, \cdots, x_n\}$
        \ENSURE $Y^2$
        \STATE $Y^1 \gets \textbf{F}(X)$
        \STATE $Y^2 \gets \textbf{F}(Y_1)$
        \STATE $S \gets \emptyset$
        \FOR{i=0 to n}
            \IF{$x_i \neq y^1_i ~\textbf{and}~ y^1_i \neq y^2_i ~\textbf{and}~ x_i \neq y^2_i$}
                \STATE $S \gets S \cup \{1\}$
            \ENDIF
            \IF{$x_i \neq y^1_i ~\textbf{and}~ x_i = y^2_i$}
                \STATE $S \gets S \cup \{1\}$
            \ENDIF
            \IF{$x_i = y^1_i ~\textbf{and}~ y^1_i = y^2_i$}
                \STATE $S \gets S \cup \{2\}$
            \ENDIF
            \IF{$x_i \neq y^1_i ~\textbf{and}~ y^1_i = y^2_i$}
                \STATE $S \gets S \cup \{3\}$
            \ENDIF
            \IF{$x_i = y^1_i ~\textbf{and}~ y^1_i \neq y^2_i$}
                \STATE $S \gets S \cup \{4\}$
            \ENDIF
        \ENDFOR
        \FOR{i=0 to n}
            \IF{$S_i = 1$}
                \STATE $y^2_i \gets x_i$
            \ENDIF
            \IF{$S_i = 4 ~\textbf{and}~ S_{i-1} \neq 3 ~\textbf{and}~ S_{i+1} \neq 3$}
                \STATE $y^2_i \gets x_i$
            \ENDIF
        \ENDFOR
    \end{algorithmic}
\end{algorithm}

\section{Experiment}

\begin{table*}[ht]
\centering
\caption{Detailed Information on the Domain-Specific Dataset}
\label{tab1}
\resizebox{0.8\textwidth}{!}{ 
\begin{tabular}{cc|ccccc}
\hline
\textbf{Dataset}&\textbf{Domain}&\textbf{Sentence}&\textbf{Correct}&\textbf{Incorrect}&\textbf{Average  Length}&\textbf{Error}\\\hline
MCSC-Train & Medical & 157193 & 78592 & 78601 & 10.9 & 146503 \\\hline
MCSC-Dev & Medical & 19652 & 9826 & 9826 & 10.9 & 18357 \\\hline
MCSC-Test & Medical & 19650 & 9825 & 9825 & 10.9 & 18286 \\\hline
EC-LAW & Legal & 2460 & 1146 & 1314 & 30.5 & 2071 \\\hline
EC-MED & Medical & 3500 & 1801 & 1699 & 50.1 & 2616 \\\hline
EC-ODW & Official Documents & 2228 & 971 & 1257 & 41.1 & 1985 \\\hline
\end{tabular}
}
\end{table*}

This paper introduces a plug-in extension module designed for Chinese spelling correction in specific domains. To evaluate the correction performance of the proposed extension module within specialized fields, we selected two domain-specific datasets for experimental validation: the MCSCSet \cite{jiang13} dataset and the EC \cite{Lv14} dataset. Detailed information on these datasets is presented in Table \ref{tab1}. The MCSC dataset comprises a training set, a validation set, and a test set, which allows for comprehensive model training and evaluation. In contrast, the EC dataset provides only a test set, primarily used for assessing the model's real-world correction effectiveness. These datasets encompass three specialized domains—medical, legal, and official documents—providing a diverse range of testing scenarios to validate the model's performance. Through these experiments, this paper analyzes the improvement in correction capability for domain-specific terms achieved by the extension module.

\subsection{Experimental Setup}

The experiments were conducted on datasets from the medical, legal, and official document domains, necessitating the creation of independent extension modules for each domain. For the medical field, we used MCSC-Train as the training set, MCSC-Dev as the validation set, and both MCSC-Test and EC-MED as the test sets. For the legal and official document fields, the model utilizes domain-specific lexicons to construct pseudo datasets. During the training of the extension modules, only the weights of the extension module are updated, while the weights of the original model remain unchanged to avoid compromising its correction capabilities. The Adam optimizer was employed, with an initial learning rate set to 5e-5. The learning rate adjustment strategy is defined by Equation \ref{eq3}:

\begin{equation}
    \label{eq3}
    \text{lr}^{(epoch)} = \text{lr}_{base} \times \delta^{\text{epoch}-1}
\end{equation}

where $\delta$ is the decay factor, set to 0.9. epoch denotes the number of training iterations, $lr_{base}$ represents the initial learning rate.

\subsection{Analysis of Experimental Results}

% \begin{table*}[htbp]
% \centering
% \caption{The model's performance on datasets from various domains.}
% \label{ta2}
% \resizebox{0.8\textwidth}{!}{%
% \begin{tabular}{|c|c|ccc|ccc|}
%     \hline
%     \multicolumn{1}{|c|}{\multirow{2}[4]{*}{\textbf{Dataset}}} & \multirow{2}[4]{*}{\textbf{Method}} & \multicolumn{3}{c|}{\textbf{Detection (\%)}} & \multicolumn{3}{c|}{\textbf{Correction (\%)}} \bigstrut\\
% \cline{3-8}          &       & \textbf{Pre} & \textbf{Rec} & \textbf{F1} & \textbf{Pre} & \textbf{Rec} & \textbf{F1} \bigstrut\\
%     \hline
%     \multirow{2}[4]{*}{MCSC-Test} & MM-BERT & 23.6  & 26.3  & 24.9  & 21.8  & 24.1  & 22.9  \bigstrut\\
% \cline{2-8}          & +Medical Plugin & 68.3  & 75.3  & 71.6  & 65.7  & 72.7  & 69.0  \bigstrut\\
%     \hline
%     \hline
%     \multirow{2}[4]{*}{EC-Law} & MM-BERT & 28.5  & 29.6  & 29.0  & 24.8  & 24.1  & 24.4  \bigstrut\\
% \cline{2-8}          & +Legal Plugin & 58.3  & 57.3  & 57.8  & 56.3  & 54.2  & 55.2  \bigstrut\\
%     \hline
%     \hline
%     \multirow{2}[4]{*}{EC-Med} & MM-BERT & 29.5  & 21.6  & 24.9  & 25.6  & 19.7  & 22.3  \bigstrut\\
% \cline{2-8}          & +Medical Plugin & 62.5  & 67.9  & 65.1  & 58.8  & 64.0  & 61.3  \bigstrut\\
%     \hline
%     \hline
%     \multirow{2}[4]{*}{EC-Odw} & MM-BERT & 35.3  & 41.7  & 38.2  & 31.0  & 36.3  & 33.4  \bigstrut\\
% \cline{2-8}          & +Official Document Plugin & 51.4  & 58.4  & 54.7  & 48.5  & 55.5  & 51.8  \bigstrut\\
%     \hline
% \end{tabular}}
% \end{table*}

% \usepackage{multirow}
\begin{table}[ht]
\centering
\caption{The model's performance on datasets from various domains.}
\label{ta2}
\begin{tabular}{cccccccc}
\hline
\multirow{2}{*}{Dataset}   & \multirow{2}{*}{Method}   & \multicolumn{3}{c}{Detection} & \multicolumn{3}{c}{Correction} \\ \cline{3-8} 
                           &                           & Pre      & Rec      & F1      & Pre      & Rec      & F1       \\ \hline
\multirow{2}{*}{MCSC-Test} & MM-BERT                   & 23.6     & 26.3     & 24.9    & 21.8     & 24.1     & 22.9     \\ \cline{2-8} 
                           & +Medical Plugin           & 68.3     & 75.3     & 71.6    & 65.7     & 72.7     & 69.0     \\ \hline
\multirow{2}{*}{EC-Law}    & MM-BERT                   & 28.5     & 29.6     & 29.0    & 24.8     & 24.1     & 24.4     \\ \cline{2-8} 
                           & +Legal Plugin             & 58.3     & 57.3     & 57.8    & 56.3     & 54.2     & 55.2     \\ \hline
\multirow{2}{*}{EC-Med}    & MM-BERT                   & 29.5     & 21.6     & 24.9    & 25.6     & 19.7     & 22.3     \\ \cline{2-8} 
                           & +Medical Plugin           & 62.5     & 67.9     & 65.1    & 58.8     & 64.0     & 61.3     \\ \hline
\multirow{2}{*}{EC-Odw}    & MM-BERT                   & 35.3     & 41.7     & 38.2    & 31.0     & 36.3     & 33.4     \\ \cline{2-8} 
                           & +Official Document Plugin & 51.4     & 58.4     & 54.7    & 48.5     & 55.5     & 51.8     \\ \hline
\end{tabular}
\end{table}

Traditional Chinese spelling correction models are generally trained on public domain datasets, limiting their effectiveness to text within common domains. However, in real-world applications, users may input text containing a substantial number of specialized terms from specific fields. These terms are often absent from the training data, leading to incorrect spell correction by the model. To address this issue, this paper proposes an extension module that learns specialized domain-specific terminology, providing relevant feature information to the correction module and thereby enhancing its ability to correct text within specialized domains.

We constructed and trained extension modules tailored to the medical, legal, and official document domains. When the original model was integrated with its respective extension modules, performance on the corresponding domain test datasets improved significantly. Table \ref{ta2} illustrates these results. It is evident that, without the extension module, the model's correction capabilities for domain-specific datasets are limited due to a lack of specialized knowledge, and it could only correct a small fraction of erroneous sentences. However, after integrating the appropriate extension module, the model's performance saw a substantial increase, with F1 scores improving by a factor of 2 to 3.

\subsection{ Case Study Analysis}

The extension module provides specialized vocabulary information from the target domain, enhancing the model's ability to accurately correct previously unseen terms without compromising its original correction capabilities. Table \ref{tab3} presents specific correction examples of the model integrated with the medical domain extension module. It can be observed that, without the extension module, the model attempts to modify incorrect characters but fails to produce correct results. In contrast, with the medical extension module enabled, the model successfully corrects the term "硝普钠" to its correct specialized term, demonstrating a significant improvement in handling domain-specific terminology. Additionally, when an unsuitable extension module was used, the model refrained from making any modifications. This behavior is attributed to the introduction of random character padding during training data construction, which trained the model to "replicate unfamiliar characters without modification." This design ensures that the model can avoid ineffective or erroneous corrections when encountering unfamiliar terms.

\begin{table}[ht]
\centering
\caption{Model Correction Result Example}
\label{tab3}
\resizebox{0.98\linewidth}{!}{ 
\begin{tabular}{c|c}
\hline
 原句子 & \textcolor{red}{肖}普\textcolor{red}{呐}的作用及副作用 \\\hline
 目标句子 & \textcolor{blue}{硝}普\textcolor{blue}{钠}的作用及副作用 \\\hline
 纠错结果（无扩展模块）& \textcolor{red}{肖}普\textcolor{red}{那}的作用及副作用 \\\hline
 纠错结果（医疗扩展模块） & \textcolor{blue}{硝}普\textcolor{blue}{钠}的作用及副作用 \\\hline
 纠错结果（法律扩展模块） & \textcolor{red}{肖}普\textcolor{red}{呐}的作用及副作用 \\\hline
\end{tabular}}
\end{table} 

\section{Conclusion}
To address the limitations of existing models in correcting errors within specialized domains, this paper introduces a plug-in extension module designed to enhance the model’s ability to correct domain-specific text. By learning specialized terminology, the extension module equips the model with domain knowledge without compromising its existing correction capabilities, enabling effective correction of specialized terms. This approach significantly improves the model's performance across multiple domain-specific datasets by incorporating the extension module, thereby effectively overcoming the model’s weaknesses in domain-specific spelling correction.
%Bibliography
\bibliographystyle{unsrt}  
\bibliography{templateArxiv}  

\end{CJK}

\end{document}